# Skeleton Focused Human Activity Recognition in RGB Video

Bruce X. B. Yu
Department of Computing
The Hong Kong Polytechnic University
Hong Kong, China
bruce.xb.yu@connect.polyu.hk

Yan Liu
Department of Computing
The Hong Kong Polytechnic University
Hong Kong, China
csyliu@comp.polyu.edu.hk

Keith C. C. Chan
Department of Computing
The Hong Kong Polytechnic University
Hong Kong, China
keith.chan@polyu.edu.hk

*Abstract*—The data-driven approach that learns an optimal representation of vision features like skeleton frames or RGB videos is currently a dominant paradigm for activity recognition. While great improvements have been achieved from existing single modal approaches with increasingly larger datasets, the fusion of various data modalities at the feature level has seldom been attempted. In this paper, we propose a multimodal feature fusion model that utilizes both skeleton and RGB modalities to infer human activity. The objective is to improve the activity recognition accuracy by effectively utilizing the mutual complemental information among different data modalities. For the skeleton modality, we propose to use a graph convolutional subnetwork to learn the skeleton representation. Whereas for the RGB modality, we will use the spatial-temporal region of interest from RGB videos and take the attention features from the skeleton modality to guide the learning process. The model could be either individually or uniformly trained by the back-propagation algorithm in an end-to-end manner. The experimental results for the NTU-RGB+D and Northwestern-UCLA Multiview datasets achieved state-of-the-art performance, which indicates that the proposed skeleton-driven attention mechanism for the RGB modality increases the mutual communication between different data modalities and brings more discriminative features for inferring human activities.

*Keywords— human activity recognition, multimodal feature fusion, attention mechanism*

I. INTRODUCTION

Human activity recognition (HAR) is one of the active fields in computer vision that facilitates many practical applications like healthcare and physical rehabilitation, interactive entertainment, and video understanding. Recently, it has undergone great improvements in single modal solutions like skeleton based methods [1], [2] and RGB video based methods [3], [4]. Given the availability of large training samples to avoid over-fitting for data-driven models, existing single modal methods also work on the direction of improving the representation power of neural network models. Skeleton based methods like ST-GCN [1] and AGC-LSTM [2] usually focus on learning spatial-temporal features to infer activity classes. Similarly, methods with RGB video inputs like C3D [5] and I3D [3] and are also modelling representations of videos to extract spatial-temporal features from the optical flow and the region of interest that are transformed or preprocessed, respectively, from the RGB video inputs.

The major limitation of RGB video is the absence of 3D structure, while skeleton sequences are also limited by the absence of texture and appearance features. As Fig. 1 shows, activities with similar motion from the NTU-RGB+D dataset [6] like writing and reading, typing and writing, pointing to something and patting other's back could hardly be recognized by ST-GCN [1]. Although there are some attempts toward vision based multimodal HAR [7-9], and even multiple sensor multimodal solutions [10-12], sensor fusion or data fusion among different data modalities remains an open problem. Specifically, as [13] summarized, it is known as co-learning where knowledge from one data modality helps with modelling in another data modality. With the advance in human body skeleton detection technology, skeleton features could be affordably and easily retrieved, which provide an extra spatial modality upon the existing RGB or depth sensors. The goal of this paper is to involve prior knowledge to the data fusion of skeleton and RGB modalities so as to improve the classification of human activities.

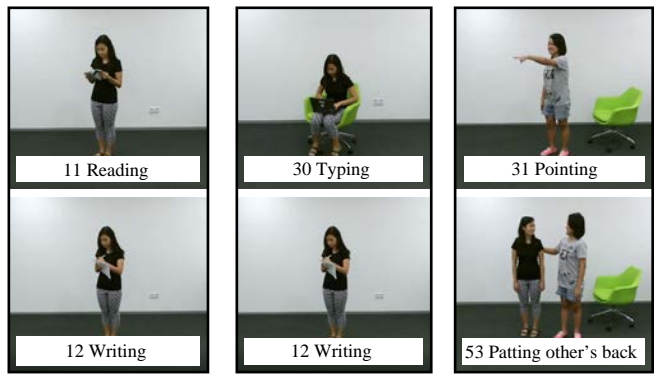

Fig. 1. Difficult action pairs like actions 11 and 12, actions 12 and 30, and actions 31 and 53 in NTU RGB+D dataset that confuse ST-GCN model

We proposed a multimodal data fusion method based on the assumption that an activity could be recognized for a human being when sufficient spatial and appearance information is satisfied in a temporal manner. Intuitively, to enable machines to have such a level of activity recognition ability, this requirement should also be satisfied and properly modelled. When we obverse a human activity, we will have a general idea about what a person is doing from the skeleton feature. However, human activities usually involve interactions with objects and other human beings. With the appearance information of objects, the searching space could be narrowed down, and this makes the recognition simpler for machines. Hence, object recognition is utilized in the jobs of [7] and [14]. With concern of what objects a person is interacting with, we focus on the body areas like head, hands and feet, where usually appears appearance information of objects. The relationship between the object and the person is evolving during the performance of an activity, hence we utilized such RGB information throughout the video by constructing a spatial temporal region of interest (ST-ROI) map for activity recognition. This strategy will alleviate the challenges from video based HAR caused by the huge data volume and diversified background appearance of human activities.

With the RGB video modality at hand, we found that deep learning methods are easy to get overfitted on existing datasets although they are claimed to be as big in volume. From our observation, the volume of the existing benchmark dataset NTU RGB+D is large, with over 50 thousand samples, but far from enough to (feed up) satisfy deep learning (DL) models like VGG nets [15] and ResNet [16] that are designed for huge datasets like ImageNet with 10 million training samples. Practices to avoid over-fitting could be by reducing the layers of DL models or adding dropout layers, which are common alternative methods to the traditional L1 and L2 regularization methods. When collecting more data is expensive, image data augmentation methods like rotation, horizontal flip, and random erasing are available but, to the best of our knowledge, none of them have been proved to improve the recognition of human activity, as it compares to the skeleton modality. Hence, we propose to utilize the features from the skeleton modality to further denoise the RGB video modality to alleviate the over-fitting issue. Precisely, we design a DL architecture that derives an attention mask from the skeleton modality for the ST-ROI of the video modality, which could be jointly trained to boost the activity recognition accuracy.

The contribution of this paper could be threefold. First, we introduce a multimodal DL architecture that fuses different data modality at the feature level. Second, for the attention mechanism, we compare fixed attention and soft attention methods. Third, our method improves the recognition of human activity on two benchmark datasets namely NTU-RGB+D and that of Northwestern-UCLA Multiview.

The organization of this paper is as follows. Section II explores the related work. Section III introduces the proposed DL architecture of our skeleton-driven attention for RGB video modality. Section IV provides the experimental results on benchmark datasets, and section V concludes the paper.

## II. RELATED WORK

HAR has experienced great progress from single sensor modalities like video based HAR [17], ambient sensors [18] and wearable sensors [19] to the paradigm of multimodal methods. With the popularity of vision based HAR, in this section, we discuss the related work of single modal HAR and multimodal HAR that focus on vision sensors.

### A. Single Modal HAR

*1) Skeleton based HAR:* Since vision sensors like motion capture (Mocap) systems, (companies like OptiTrack, Qualisys, and VICON), depth cameras like Kinect and RealSense and even RGB cameras have the ability to detect human body skeletons, skeleton based activity recognition is a booming area in computer vision. Traditionally, algorithms for skeleton based HAR focus on the sequential character of an activity. Hence, traditional algorithms like SVM, HMM, DTW are commonly used [20], which is then dominated by DL algorithms [21]. Wang et al. [21] reviewed DL models for HAR tasks, which includes Deep neural network (DNN), Convolutional Neural Network (ConvNets, or CNN), Stacked autoencoder (SAE), etc. Existing researches of skeleton based HAR mainly focus on three directions for the improvement of activity recognition. The first direction focuses on data preprocessing and data cleaning. For example, Liu et al. [22] proposed a method that removes the skeleton joint noise by learning a model that reconstructs more accurate skeleton data. Similar jobs have been proposed by Pengfei et al. [23]. The second approach improves the HAR benchmarks by proposing novel learning or representing models. Liu et al. [24] proposed a context aware LSTM model that could learn which part of joints contribute to the HAR. Since the induction of ST-GCN [1] some enhanced versions of Graph Convolutional Network (GCN) models have been proposed that improve the ST-GCN by considering other physical prior knowledge. For example, Li et al. [25] tries to model discriminative features from actional and structural links of the skeleton graph. Except GCN, motivated by cooccurrence learning, Chao et al. [26] proposed the hierarchical cooccurrence network (HCN) that learns point-level features aggregated to cooccurrence features with a hierarchical methodology. The co-occurrence features refer to the interactions and combinations of some subsets of skeleton joints that characterizes an action [27]. Considering both the graph and cooccurrence characteristics, Si et al. [2] proposed AGC-LSTM that achieved high accuracy on the NTU-RGB-D dataset. However, by the time of this job, the directed graph network (DGN) [28] achieved the state-of-the-art performance on the NTU RGB+D dataset with a smaller margin than the AGC-LSTM.

*2) Video based HAR:* Since video data are relatively easier to get, big datasets like UCF-101 and HMDB-51 and Kinetics are collected as benchmarks in the video based HAR area. Carreira proposed I3D [3] that uses pre-trained inflated Inception-V1 mode miniKinetics as back bone to improve the performance of UCF-101 and HMDB-51 by end-to-end fine-tuning. Two data modalities RGB and optical flow, extracted by the TV-L1 algorithm are used, as in their two-stream model. It turns out that the optical flow modality performs better for UCF-101 and HMDB-51 dataset but is surpassed by RGB modality on miniKinetics. On top of I3D, Xie et al. [4] considers about the speed-accuracy trade-offs in video classification, and proposed a S3D model that further improves the performance of [29]. It is worth mentioning that the S3D [4] was implemented on 56 GPUs with a batch size set to 6 per GPU, which reflects the huge computational cost of video based HAR. It is intuitive that fusing the results of S3D with skeleton

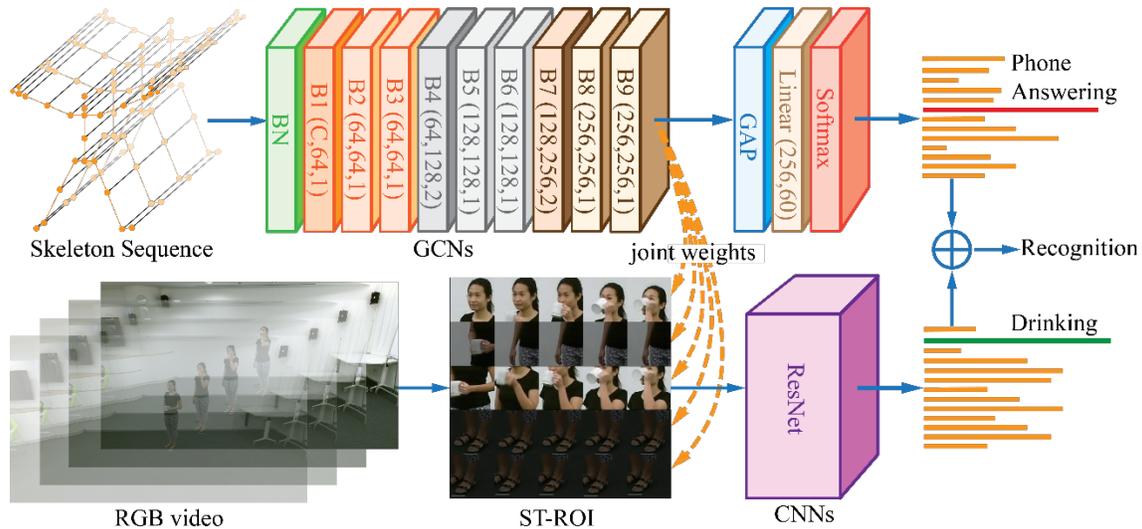
Fig. 2. Deep learning architecture of skeleton-driven attention for RGB video modality

based methods could boost the recognition accuracy. However, from our observation, the S3D will stick at around 12% for the Top-1 accuracy for NTU RGB+D. This might be due to the fact that S3D is designed for datasets like UCF-101 and Kinetics that are different with the NTU RGB+D. Precisely, activities in UCF-101 and Kinetics cover broader activity categories of both indoor and outdoor. The indoor activities in NTU RGB+D might be more challenging for such video-based methods. Another reason might be due to the limitation of computational resources as a workstation with 56 GPUs is not common.

*B. Multimodal HAR*

It has been generally accepted that multimodal HAR approaches have the potential to improve recognition and could be capable of distinguishing difficult activities. The multimodal fusion analysis of [12] for the Opportunity dataset indicates that feeding more data channels to its proposed DeepConvLSTM would deliver improved performance. Similarly, the experimental results of [30] for NTU RGB+D 120 dataset also indicate that extra data modalities contribute to classification accuracy. Multimodal HAR could be roughly categorized into two classes: vision-based multimodal [7-9] and vision-wearable-based multimodal [10-12]. Algorithms for multimodal HAR share the similar trend that uses DL to extract discriminative features. Baltrušaitis et al. [13] summarized five technical challenges of multimodal solutions: representation, translation, alignment, fusion, and co-learning. The key issue of multimodal methods is to find proper ways of data fusion with the co-learning concept in mind.

The 4DHOI model proposed by Wei et al. [7] attempts to represent both 3D human poses and contextual objects in events by using a hierarchical spatial temporal graph. The fusion concept of [9] has two approaches namely late fusion and intermediate fusion. The late approach simply combines the emission probabilities from two modalities. In the intermediate fusion scheme, the skeleton and RGB modalities are first pretrained separately, then their high-level representations are concatenated to generate a shared representation. Pan et al. [31] proposed a cross-stream selective network (CSN) that leverages the correlation and complementarity of different input streams. CSN is designed to select the most discriminative temporal frames aligned to spatial frames and globally endows different weights to RGB and optical flow groups. Unlike CSN, our method will select which body part from the RGB stream will provide extra discriminative information. This concept has been attempted by [32], which achieved decent improvement by using an attention mechanism that focuses on two hands. However, the method of [32] might neglect some activities with human-object interaction that involves lower body like "put on shoes" and "take off shoes". Other similar jobs also have been proposed in [33] and [34]. In our method, we utilize the effective attention features from the skeleton modality that focus on more body areas including the head and feet instead of just two hands as in some existing jobs [32-34].

### III. OUR PROPOSED METHOD

We have two sets of features extracted from different data modalities (skeleton and RGB signals) as input of our multimodal HAR method. State-of-the-art spatial temporal graph convolutional models like ST-GCN [1], AGC-LSTM [2], and DGN [28] could learn effective representation from the skeleton modality that has spatial importance for different skeleton joints. Meanwhile, models like I3D [3] and S3D [4] have the potential to learn discriminative features directly from video inputs but require huge computational resources. On the other hand, DL models like VGG nets [15] and ResNet [16] are effective to gain RGB features from images but are usually encountered with overfitting for datasets that are not big enough. Consequently, fusing the complementary skeleton and RGB features could be beneficial for action recognition.

There are various methods available for feature fusion. The choice of fusion strategy should rely on the characteristics of the data modalities. Most existing fusion is either at the decision level or at later layer concatenation, which is a lack of considering the borrowing of features from one modality to improve the performance of another modality, and eventually improves the performance. Since from the data of the RGB modality, it could be easy to make DL models with high representation power to overfit to the background noise. We propose a multimodal deep learning architecture (see Fig. 2) that borrows spatial knowledge from the skeleton modality to alleviate this issue. On the other hand, the lack of appearance features from the skeleton modality renders it hard to distinguish the activities, especially those with object interactions. Hence, a proper feature representation of RGB modality contributes back to the skeleton modality and boosts the ultimate performance.

In this section, we first introduce our multimodal DL architecture by first describing the subnetworks utilized to learn features from the RGB and skeleton modalities and then elaborating the feature fusion mechanism between the two modalities.

### A. Construct ST-ROI from RGB Modality

Intuitively, video based models like I3D [3] and S3D [4] could be the first choice to learn discriminative features from the RGB modality. However, these models require huge computational resources of RAM and GPU memory and will take longer to converge. From our observation, even with pre-training, those models could not converge well with some datasets. Hence, we propose to build a spatial temporal ROI from the RGB modality and use general CNN models to retrieve effective features. Unlike the method proposed in [32-34] that focus on the appearance features of two hands, we build the spatial region of interest (ROI) that focuses on body parts including head, two hands and two feet in a temporal manner.

Let us notate $V = \{V^{(i)} | i = 1,2,...,N\}$ as the RGB modality that has $N$ video samples for training. Then an ordered video sequence of an activity in the time interval $[1, T]$ could be represented as $V_i = [f_1^{(i)}, ..., f_t^{(i)}, ..., f_T^{(i)}]$, where $f_t^{(i)}$ is the frame at time $t$. To crop the spatial ROI from an activity video, we use joints of the skeleton retrieved with the OpenPose algorithm introduced in [35], which is relatively more accurate than the skeleton of the Kinect v2 sensor. Given an RGB frame $f_t^{(i)}$, we define the spatial transformation function as

$$R_{tj}^{(i)} = g(f_t^{(i)}, o_{tj}^{(i)}), j \in (1, 2, ..., K), K \leq M \quad (1)$$

where $R_{tj}^{(i)}$ and $o_{tj}^{(i)}$ are the jth joint of the spatial ROI and the jth joint of the OpenPose skeleton at time $t$, respectively. $K$ is the index of the skeleton joints, which is equal or smaller than the total number of the skeleton joints $M$. Given $V^{(i)} = [f_1^{(i)}, ..., f_t^{(i)}, ..., f_T^{(i)}]$, we then conduct a temporal sampling that selects $L$ representative frames at time $\tau = \{\tau + interval \times l | l = 1, ..., L, interval = T/L\}$ and concatenate them into a square ST-ROI as shown in the one subject case of Fig. 3. For activities that have two subjects, we crop the ST-ROIs of both subjects as shown in the two subjects case of Fig. 3. The ST-ROI significantly reduces the data volume of the RGB video modality and still preserves the object information of activities. The sub temporal ROI at time $\tau$ will have $K$ sub spatial ROIs, which could be vertically concatenated and represented as $R_\tau^{(i)}$. On the other hand, the sub spatial ROI of the jth joint will have $L$ sub temporal ROIs and could be horizontally concatenated and represented as $R_j^{(i)}$. The ST-ROI of $V^{(i)}$ could then be notated as $R^{(i)}$ that contains $K \times L$ sub ST-ROIs $R_{\tau j}^{(i)}$.

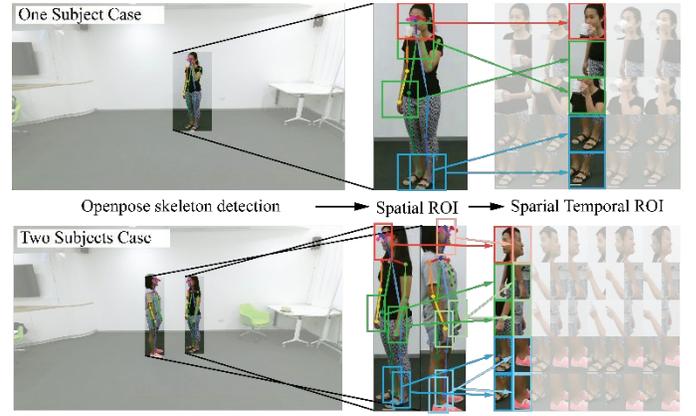

Fig. 3. Process of constructing the Spatial Temporal ROI

### B. Learn Joint Weights from the Skeleton Modality Units

For the skeleton modality, given a set of $M$ joints in a skeleton frame observed at time t, let us represent it as $j_t = (j_{t1}, ..., j_{t2}, ..., j_{tM})$. The $i$-th training sample that starts at time $t = 1$ and ends at time $T$ with skeleton frames collected at regular intervals can, therefore, be represented as a sequence of $T$ skeleton frames, $J^{(i)} = [j_1, j_2, ..., j_t, ..., j_T]$. With a total of $N$ training samples, the skeleton modality and video modality of a dataset could be represented as $J = \{J^{(i)} | i = 1,2,...,N\}$. We adopt a spatiotemporal graph to model the spatial and temporal structure of $J^{(i)}$. The structure of the Graph Convolutional Network (GCN) follows [1]. Fig. 4 illustrates an example of the constructed spatiotemporal skeleton graph, where the joints are represented as vertices and their natural connections are represented as spatial edges (the orange lines in Fig. 4, left). For the temporal dimension, the corresponding joints between two adjacent skeleton frames are connected with temporal edges (the black lines in Fig. 4, left). The attribute of a vertex is the corresponding 3D coordinate values of each joint. The skeleton graph at time $t$ could be symbolized as $\vartheta_t = \{v_t, \varepsilon_t\}$, where $v_t$ denotes the skeleton joints and $\varepsilon_t$ denotes the skeleton bones, respectively. In this graph, the node set $v = \{v_{tj} | v_{tj} = j_{tj}, t = 1, ..., T, j = 1, ..., M\}$ contains all joints of a skeleton sequence.

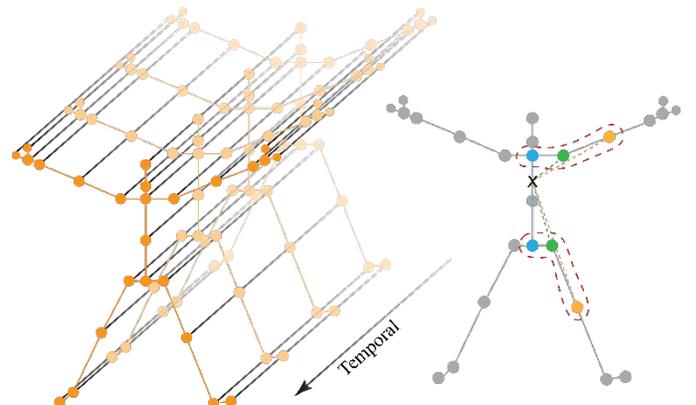

Fig. 4. (left). The structure of a spatiotemporal graph. (right). Illustration of the spatial mapping strategy. Different colors denote different subsets. Green denotes the vertex itself; Yellow denotes the farther centrifugal subset; Blue denotes the closer centripetal subset.

## C. Graph Convolutional Operation

To represent the sampling area of convolutional operations, a neighbor set of a node $v_{ti}$ is defined as $N(v_{ti}) = \{v_{tj}|d(v_{ti}, v_{tj}) \leq D\}$, where D is the minimum path length of $d(v_{ti}, v_{tj})$. The right sketch in Fig. 4 shows this strategy, where × represents the center of gravity of the skeleton. The sampling area $N(v_{ti})$ is enclosed by the curve. In detail, the strategy empirically uses 3 spatial subsets: the vertex itself (the green circle in Fig. 4, right); the centripetal subset, which contains the neighboring vertices that are closer to the center of gravity (the blue circle); and the centrifugal subset, which contains the neighboring vertices that are farther from the gravity center (the yellow circle). Suppose there is a fixed number of $K$ subsets in the $N(v_{ti})$, every neighbor set will be labelled numerically with a mapping $l_{ti}: N(v_{ti}) \to \{0, ..., K-1\}$. Temporally, the neighborhood concept is extended to temporally connected joints as $N(v_{ti}) = \{v_{qj}|d(v_{tj}, v_{ti}) \leq K, |q-t| \leq \Gamma/2\}$, where $\Gamma$ is the temporal kernel size that controls the temporal range of the neighbor set. Then the graph convolution could be computed as

$$Y_{output}(v_{ti}) = \sum_{v_{tj} \in N(v_{ti})} \frac{1}{Z_{ti}(v_{tj})} f_{in}^t(v_{tj}) W(l(v_{tj})) \quad (2)$$

where $f_{in}^t(v_{tj})$ is the feature map that gets the attribute vector of $v_{tj}$, $W(l(v_{tj}))$ is a weight function $W(v_{ti}, v_{tj}): N(v_{ti}) \to R^c$ that could be implemented by indexing a tensor of $(c, K)$ dimension. $Z_{ti}(v_{tj}) = |\{v_{tk}|l_{ti}(v_{tk}) = l_{ti}(v_{tj})\}|$ is a normalization term that equals the cardinality of the corresponding subset.

## D. Joint Weights

With implementation of graph convolution on the skeleton modality, the output of each vertex on the graph could be used to infer the importance of the corresponding skeleton joint. The feature map of the skeleton sequence could be represented by a tensor of $(C, T, V)$ dimensions, where $V$ denotes the number of vertices, $T$ denotes the temporal length and $C$ denotes the number of attributes of the joint vertex. With the specific partitioning strategy determined, it could be represented by an adjacent matrix **A** with its elements indicating if a vertex $v_{tj}$ belongs to a subset of $N(v_{ti})$. The graph convolution is implemented by performing a $1 \times \Gamma$ classical 2D convolution and multiplies the resulting tensor with the normalized adjacency matrix $\Lambda^{-\frac{1}{2}} A \Lambda^{-\frac{1}{2}}$ on the second dimension. With $K$ partitioning strategies $\sum_{k=1}^{K} A_k$, equation 2 could be transformed into

$$Y_{feature} = \sum_{k=1}^{K} \Lambda_k^{-\frac{1}{2}} A_k \Lambda_k^{-\frac{1}{2}} f_{in} W_k \odot M_k \quad (3)$$

where $\Lambda_k^{ii} = \sum_j(A_k^{ij}) + \alpha$ is a diagonal matrix with $\alpha$ set to 0.001 to avoid empty rows. $W_k$ is a weight tensor of the $1 \times 1$ convolutional operation with $(C_{in}, C_{out}, 1, 1)$ dimensions, which represents the weighting function of equation (2). $M_k$ is an attention map with the same size of $A_k$, which indicates the importance of each vertex. $\odot$ denotes the element-wise product between two matrixes. $Y_{feature}$ is a tensor with the size of $(c, t, v)$ with $c$ as the number of output channels, $t$ as the temporal length and $v$ as the number of vertices, which could be used to infer the activity class and transformed as joint weights to provide knowledge for the RGB modality. The joint weights that represent the importance of joints could be interpreted as

$$J_{weight} = \frac{1}{tc} \sum_1^t \sum_1^c \sqrt{Y_{feature}^2} \quad (4)$$

where $t$ and $c$ are the output dimensions of the convolutional graph that represents the temporal length and out channel, respectively. $J_{weight}$ is a vector that contains weights for different skeleton joints.

## E. Joint Weights

We propose a spatial weight mechanism for the RGB frames to enable the machine being capable to focus on RGB features that will provide discriminative information. In an explainable way, this will make the machine more capable as it intuitively mimics activity recognition of human ways. Researchers also attempt to learn an attention weight from the RGB modality itself. From the result of [36] that has tested four variants of attention mechanism on the job of Convolutional LSTM [37], there are very few or even no performance improvements, although it decreases the model size. Moreover, the gesture datasets used by [36] have very consistent backgrounds, which might make the attention mechanism even less effective for datasets that have varying complex backgrounds. Hence, we do not continue to explore the contribution of the attention mechanism in this job. Instead, we use the joint weights from the skeleton modality and multiply it with the ST-ROI to reduce the noise of the RGB modality. The weighted ST-ROI of the $i$th training sample $R'^{(i)}$ could be mapped from $R^{(i)}$ by using a function

$$R'^{(i)} = h(R_j^{(i)}, J_{weight\_j}), j \in (1, 2, ..., K) \quad (5)$$

where $J_{weight\_j}$ is the joint weight of the $j$th joint and $R_j^{(i)}$ is the sub spatial ROI of the $j$th joint. Fig. 5 shows the process of function (5) that denoises the RGB modality.

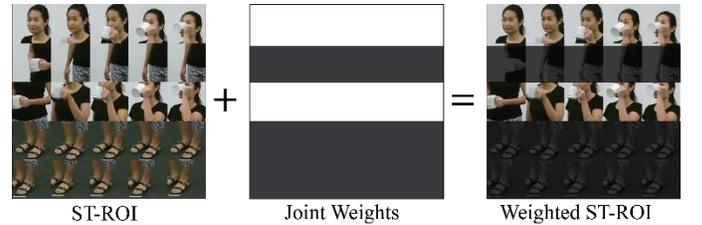

ST-ROI     Joint Weights     Weighted ST-ROI

Fig. 5. Multiply joint weights to the ST-ROI

## F. Objective Function

We build the end-to-end format with the sum of a collection of loss terms from different data modalities that are supervised by the activity label, which are explained below:

$$\mathcal{L} = \frac{1}{N} \sum_{i=1}^{N} \left(\hat{y}^{J(i)} - y^{(i)}\right)^2 + \frac{1}{N} \sum_{i=1}^{N} \left(\hat{y}^{R(i)} - y^{(i)}\right)^2 \quad (6)$$

where $\hat{y}^{J(i)}$ is the output of the skeleton modality from GCNs by using equation (3), and $\hat{y}^{R(i)}$ is the output of the RGB modality from the ResNet [16].

*G. Training and Optimization*

Given the objective function, there are many deep learning multimodal data fusion strategies that could be adopted to pursue high recognition accuracy. For example, in [32], both the pose prediction loss that encourages the model reserve of the pose repression during training and the pose attraction loss that makes the skeleton attention more similar to humans can lead to attempting to improve performance. To ease the process of proving the effectiveness of our multimodal method, we avoid using such fine-tuning skills and hyperparameter tuning skills and adopt a vanilla training process to verify the effectiveness of our model as we have already proposed to fuse different data modalities at a feature level. More precisely, the training steps are illustrated in Algorithm 1.

---

**Algorithm 1:** Training of the multimodal HAR network

**Input**: $V = \{V^{(i)} \mid i = 1,2,...,N\}$: RGB videos
$J = \{J^{(i)} \mid i = 1,2,...,N\}$: skeleton joint coordinates
$K$: the number of sub spatial ROIs
$L$: the number of sub temporal ROIs

1. Train GCNs with skeleton modality of joint coordinates $J$.
2. Construct a $K \times L$ ST-ROI $R^{(i)}$ from the RGB videos $V$.
3. Extract joint weights $J_{weight}$ by feeding $J$ to the trained GCNs.
4. Construct weighted ST-ROI $R'^{(i)}$ from step 2 and 3.
5. Train ResNet with $R'^{(i)}$.
6. Ensemble results from steps 1 and 5 to infer the activity class.

**Output**: learned multimodal network

---

## IV. EXPERIMENTAL ANALYSIS

In this section, we introduce the datasets used in the experimental analysis and the comparison of our method with state-of-the-art methods.

*A. Datasets*

We performed our experiments on the following two human action recognition datasets: NTU RGB+D Dataset [6] and Northwestern-UCLA Multiview Action 3D Dataset [38].

*1) NTU RGB+D dataset:* NTU RGB+D dataset [6] was collected with by Kinect v2 sensors and contains over 56K samples of 60 different activities including individual activities, interactions between multiple people, and health-related events. The activities were performed by 40 subjects and recorded from 80 viewpoints. We followed the cross-subject and cross-view split protocol from [6]. Since this dataset provides multiple modalities of data from the Kinect v2 sensor, this dataset is highly suitable for testing multimodal HAR methods.

*2) Northwestern-UCLA Multiview dataset:* The Northwestern-UCLA Multiview dataset was collected by [38], which contains more interactions between human subjects and objects. The dataset has 12 action categories with each of them performed by 10 actors. It has 1,494 samples in total, which includes 518 samples from view 1, another 509 samples from view 2 and 467 samples from view 3. We follow the evaluation method in [38] that takes samples from two camera views for training and the remaining camera view for testing. Precisely, in table 2, it is denoted as V_1,2^3 that means samples from views 1 and 2 are taken for training, and samples from the view 3 are used for testing.

*B. Implementation Details*

For the RGB modality, the height and width of sub ST-ROIs of NTU-RGB+D and Northwestern-UCLA datasets are 96 and 48 pixels, respectively. We set both K and L to 5 to construct the ST-ROI. Hence the input size for NTU-RGB+D and Northwestern-UCLA datasets are 480×480 and 240×240, respectively. The ST-ROI of both datasets were resized to 225×225 and normalized before feeding them into ResNet. As the data volume of Northwestern-UCLA is relatively small, we perform random selection to the RGB video frames and randomly flip them. We adopted ResNet18 that has 18 layers for both datasets. Both implementations used the SGD optimizer with initial learning rate set as 0.1 which is divided by 10 at the 45th and 55th epochs and ended at the 65th epoch. The minibatch size is set at 64. All experiments are conducted on a workstation with 4 GTX 1080 Ti GPUs.

*C. Results*

*1) Ablation study:* Table I and table II show several experiments with different data modalities and their ensembled results. The statistics show considerable improvements by ensemble results of the RGB modality to the results of the skeleton modality with different training methods numbered as 2, 3, and 4 on both NTU-RGB+D and Northwestern-UCLA datasets. By comparing training methods 2 and 3, we could observe that the proposed skeleton driven weights mechanism is able to effectively improve the discriminative power of the ST-ROI features. Contrasting training method 3 with that of 4, tuning the weights of GCN together with the ResNet, could be of benefit for the RGB modality. However, in terms of the overall performance improvement when aggregating the results of the RGB modality and the skeleton modality, fixing the GCN weights by set it at evaluation mode will achieve better ensemble results for both datasets. In Fig. 6, we illustrated the effectiveness of the weighted ST-ROI method that improves the recognition accuracy of every activity of Northwestern-UCLA Dataset. It also indicates that even for some activities the accuracy of the RGB modality is not as good as the skeleton modality. However, it will still contribute to the overall performance.

TABLE I. ABLATION STUDY FOR NTU RGB+D DATASET WITH CROSS-SUBJECT (CS) AND CROSS-VIEW (CV) PROTOCOL (ACCURACY IN %). ○ MEANS IN EVALUATION MODE. ○ MEANS IN EVALUATION MODE.

| # | Methods | Skeleton | RGB | CS | CV | Avg |
|---|---|---|---|---|---|---|
| 1 | GCN | √ | - | 81.4 | 88.0 | 86.7 |
| 2 | ResNet18 | - | √ | 76.4 | 80.5 | 78.5 |
| 3 | ResNet18+Weights | √ | √ | 86.8 | 89.3 | 88.1 |
| 4 | ResNet18+Weights | ○ | √ | 82.8 | 87.4 | 85.1 |
| 5 | Ensemble (1+2) | ○ | ○ | 90.1 | 93.1 | 92.3 |
| 6 | Ensemble (1+3) | ○ | ○ | 90.2 | 93.5 | 92.4 |
| 7 | Ensemble (1+4) | ○ | ○ | 91.5 | 95.0 | **93.3** |

TABLE II. ABLATION STUDY FOR THE NORTHWESTERN-UCLA MULTIVIEW DATASET WITH CROSS-VIEW SETTING (ACCURACY IN %). ○ MEANS IN EVALUATION MODE.

| # | Methods | Skeleton | RGB | $V^3_{1,2}$ |
|---|---|---|---|---|
| 1 | GCN | √ | - | 82.9 |
| 2 | ResNet18 | - | √ | 77.1 |
| 3 | ResNet18+Weights | √ | √ | 84.9 |
| 4 | ResNet18+Weights | ○ | √ | 83.2 |
| 5 | Ensemble (1+2) | ○ | ○ | 90.5 |
| 6 | Ensemble (1+3) | ○ | ○ | 90.3 |
| 7 | Ensemble (1+4) | ○ | ○ | **91.6** |

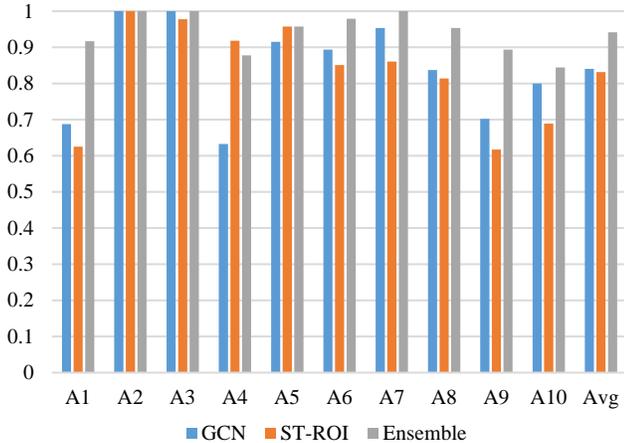

Fig. 6. Recognition accuracy improvement in every activity of Northwestern-UCLA Dataset

*2) Comparison with state-of-the-Art methods:* We show the performance comparison with other state-of-the-art methods that use the skeleton modality, EGB modality and multiple modality in table III and table IV for the NTU-RGB+D and Northwestern-UCLA datasets, respectively. Our skeleton joint-weighted ST-ROI method achieved state-of-the-art performance on both datasets with a vanilla implementation by using the basic ResNet model which is the ResNet18. Whereas, to achieve good performance, existing RGB based methods usually utilize much more complex CNN models like ResNet50 in [32].

TABLE III. RESULTS FOR THE NTU RGB+D DATASET WITH CROSS-SUBJECT (CS) AND CROSS-VIEW (CV) SETTINGS (ACCURACIES IN %)

| Methods | Skeleton | RGB | CS | CV | Avg |
|---|---|---|---|---|---|
| Lie Group [39] | √ | - | 50.1 | 52.8 | 51.5 |
| Dynamic Skeletons [40] | √ | - | 60.2 | 65.2 | 62.7 |
| Part-aware LSTM [6] | √ | - | 62.9 | 70.3 | 66.6 |
| GCA-LSTM [41] | √ | - | 74.4 | 82.8 | 78.6 |
| View-invariant [22] | √ | - | 80.0 | 87.2 | 83.6 |
| ST-GCN [1] | √ | - | 81.5 | 88.3 | 84.9 |
| DPRL+GCNN [42] | √ | - | 83.5 | 89.8 | 86.7 |
| 2S-AGCN [43] | √ | - | 88.5 | 95.1 | 91.8 |
| AGC-LSTM [2] | √ | - | 89.2 | 95.0 | 92.1 |
| DGNN [28] | √ | - | 89.9 | **96.1** | 93.0 |
| C3D [5] | - | √ | 63.5 | 70.3 | 66.9 |
| Glimpse Clouds [32] | - | √ | 86.6 | 93.2 | 89.9 |
| DSSCA - SSLM [8] | √ | √ | 74.9 | - | - |
| STA-Hands [33] | √ | √ | 82.5 | 88.6 | 85.6 |
| Hands Attention [34] | √ | √ | 84.8 | 90.6 | 87.7 |
| Our Multimodal Method | √ | √ | **91.5** | 95.0 | **93.3** |

TABLE IV. RESULTS FOR THE NORTHWESTERN-UCLA MULTIVIEW ACTION 3D DATASET WITH CROSS-VIEW SETTING (ACCURACY AS A PERCENT).

| Methods | Skeleton | RGB | $V^3_{1,2}$ |
|---|---|---|---|
| Lie Group [39] | √ | - | 73.3 |
| HBRNN-L [44] | √ | - | 78.5 |
| View-invariant [22] | √ | - | 86.1 |
| Ensemble TS-LSTM [45] | √ | - | 89.2 |
| Hankelets [46] | - | √ | 45.2 |
| nCTE [47] | - | √ | 68.6 |
| NKTM [48] | - | √ | 75.8 |
| Glimpse Clouds [32] | - | √ | 90.1 |
| Our Multimodal Method | √ | √ | **91.6** |

## V. CONCLUSION

We proposed a multimodal DL architecture for vision based HAR that utilized the data from skeleton and RGB video modalities. The method borrows the skeleton-driven attention feature from the skeleton modality and contributes the performance of the RGB modality, which in turn contributes to the ultimate performance. This method achieves state-of-the-art performance with the NTU-RGB+D and N-UCLA datasets compared to methods that use skeleton, RGB video, or both at test time. Testing results of the RGB modality by using a fixed attention mechanism is not as good as that of the soft mode but achieved better performance in the final results when ensemble with the results of the skeleton modality. In the future, we will further investigate the other aspects that affect performance of the multimodal HAR by designing architectures with more prior knowledge and make them more explainable and improvable.